\documentclass[lettersize,conference]{ieeeconf}

\IEEEoverridecommandlockouts                              
                                                          
\usepackage{cite}
\usepackage{amsmath,amssymb,amsfonts}
\usepackage{algorithmic}
\usepackage{graphicx}
\usepackage{textcomp}
\usepackage{xcolor}
\def\BibTeX{{\rm B\kern-.05em{\sc i\kern-.025em b}\kern-.08em
    T\kern-.1667em\lower.7ex\hbox{E}\kern-.125emX}}
    
\usepackage{xspace}
\let\OldTexttrademark\texttrademark
\renewcommand{\texttrademark}{\OldTexttrademark\xspace }%

\usepackage [english]{babel}
\usepackage [autostyle, english = american]{csquotes}
\MakeOuterQuote{"}
    
\usepackage{hyperref}
\newcommand{\email}[1]{\href{mailto:#1}{#1}}

\usepackage{bm}

\usepackage{cleveref}
\usepackage{afterpage}
\usepackage{gensymb}
\usepackage{paralist}
\usepackage{multirow}
\usepackage{booktabs}
\usepackage{float}

\renewcommand{\baselinestretch}{0.97} 

\begin{document}


\title{Design of Trimmed Helicoid Soft-Rigid Hybrid Robots}

\author{Zach J. Patterson$^{1,2,\dagger}$, Emily R. Sologuren$^{1}$, and Daniela Rus$^{1}$
\thanks{*This work was done with the support of National Science Foundation EFRI program under grant number 1830901 and the Gwangju Institute of Science and Technology.}
\thanks{$^{1}$ Computer Science and Artificial Intelligence Laboratory, MIT. \email{rus@csail.mit.edu}}%
\thanks{$^{2}$ Mechanical and Aerospace Engineering, Case Western Reserve University. \email{zpatt@case.edu}}%
\thanks{$^{\dagger}$This work was completed when Zach J. Patterson was at MIT. He is now at Case Western.}%
}

\maketitle

\begin{abstract}
As soft robot design matures, researchers have converged to sophisticated design paradigms to enable the development of more suitable platforms. Two such paradigms are soft-rigid hybrid robots, which utilize rigid structural materials in some aspect of the robot's design, and architectured materials, which deform based on geometric parameters as opposed to purely material ones. In this work, we combine the two design approaches, utilizing trimmed helicoid structures in series with rigid linkages. Additionally, we extend the literature on wave spring-inspired soft structures by deriving a mechanical model of the stiffness for arbitrary geometries. We present a novel manufacturing method for such structures utilizing an injection molding approach and we make available the design tool to generate 3D printed molds for arbitrary designs of this class. Finally, we produce a robot using the above methods and operate it in closed-loop demonstrations.
\end{abstract}

\section{Introduction}

While soft robotics has undergone explosive growth in the last 15 years \cite{rusDesignFabricationControl2015c}, real world impact remains somewhat limited due in part to the difficulty of producing robust and versatile hardware platforms \cite{hawkesHardQuestionsSoft2021a}. To address this problem, and inspired by the complex material compositions of the biological world, there is growing interest in the deployment of active structures that include materials across stiffness scales, from very soft to very rigid \cite{buchnerVisioncontrolledJettingComposite2023, bernSimulationFabricationSoft2022a, zhuSoftRigidHybridGripper2023, zhangGeometricConfinedPneumatic2020, coevoetPlanningSoftRigidHybrid2022, mathewSoRoSimMATLABToolbox2023}. We refer to such robots as soft-rigid hybrids. These designs have been created with the hope that they can take advantage of the precise, load bearing capability of rigid structures along with the kinematic adaptability and favorable collision robustness of soft structures \cite{pattersonDesignControlModular2024a}.

 \begin{figure}
\includegraphics[width=0.48\textwidth]{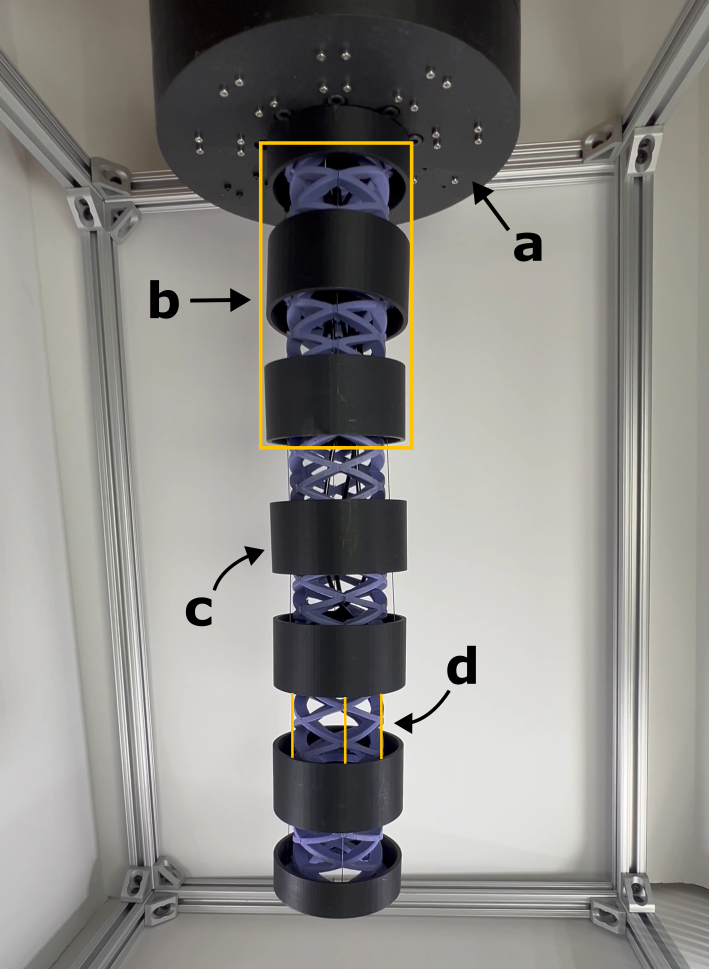}
\caption{\textbf{Soft-Rigid Helicoid Arm.} For scale, the robot is approximately 0.45 meters from base to tip. (\textbf{A}) Rotating base. (\textbf{B}) One module is composed of two helical segments. (\textbf{C}) Rigid plate that connects two segments together. (\textbf{D}) Each module (two helicoid segments) are controlled by three cables highlighted in orange. }\label{fig:robot}
\end{figure}

Another growing design trend in the literature is the use of architectured structures to allow more design flexibility \cite{bergerMechanicalMetamaterialsTheoretical2017, xiaResponsiveMaterialsArchitected2022, barthelatArchitecturedMaterialsEngineering2015}. Researchers in soft robotics have been interested in architectured structures and metamaterials for a long time, but most of this work exists at the level of fundamental research, with relatively few works deploying full robot systems necessary to produce versatile robot platforms. A recent notable exception of this is the work of Guan et al., who deploy a tendon-driven robot manipulator with a structural design based on the trimmed helicoid method \cite{guanTrimmedHelicoidsArchitectured2023a}. In short, trimmed helicoids are interlocking helices with a cylindrical cut down the radial center of the structure. They are inspired by and extremely similar to wave springs. The chief advantage of this design, as with all architectured structures, is that bulk structural properties can be designed independently of material properties. Additionally, within a robotic structure, because of the high volume to mass ratio of trimmed helicoids, they allow for higher payloads than equivalent bulk material structures. While the previous work has demonstrated the utility of the trimmed helicoid concept \cite{wanDesignAnalysisRealTime2023, skorinaSoftHybridWave2020, salgueiroHighlyIntegrated3D2021}, there are significant improvements to be made. All design oriented modeling in the previous work was based on the Finite Element Method (FEM) \cite{guanTrimmedHelicoidsArchitectured2023a}. While this is likely to provide the best agreement with reality, it is limited as a design tool as a new, expensive analysis must be performed for any potential design. Another limitation is that the previous works exclusively rely on a 3D printing workflow \cite{gunarathnaDesignCharacterization3D2022}. While this is highly useful for rapidly iterating upon lab prototypes, 3D printed elastomeric materials remain inferior to standard commercial silicone elastomers on important material properties such as viscoelasticity, tear strength and fatigue life \cite{wallin3DPrintingSoft2018a, sachyanikeneth3DPrintingMaterials2021}. 

In this work, we combine the research thrusts discussed in the previous two paragraphs. Namely, we integrate the trimmed helicoid concept into serial soft-rigid hybrid structures. In doing so, we hope to take the first steps towards a more capable general purpose soft robot manipulator. We also address a few of the aforementioned frontiers in the trimmed helicoid literature. Specifically, we develop a modeling approach based on elementary beam mechanics that allows for a simplified, fully analytical model of the compressive stiffness of arbitrary trimmed helicoid geometries. We extend this modeling approach in a semi-empirical fashion to model bending stiffness. To manufacture the robots, we injection mold instead of 3D print and provide an openly available CAD model that generates 3D printable injection molds for arbitrary trimmed helicoid geometries. Finally, we fabricate a characteristic robot and provide a few simple demonstrations of performance. An image of the resulting robot is shown in Figure \ref{fig:robot}.





In summary, we present the following contributions:
\begin{compactitem}
    \item A new design paradigm combining the trimmed helicoid with serial soft rigid hybrids;
    \item Straightforward analytic design oriented models for trimmed helicoids and SRHs;
    \item A novel manufacturing strategy for trimmed helicoids that allows for the use of industrial-quality elastomers and mass manufacturing;
    \item An open source design tool that allows instantaneous generation of the CAD for the production of a robot;
    \item Experimental characterization of the above and a proof-of-concept robot.
\end{compactitem}


\section{Design of trimmed helicoid soft-rigid hybrid structures}\label{sec:design}

We begin with a clarification of the potential benefits from using serial soft-rigid hybrids and trimmed helicoids together. Each of these two concepts brings its own rationale and strengths. The SRH paradigm allows for structures to leverage the discrete mechanics of rigid contact to actively create de-facto rigid bodies in certain configurations. This provides a form of discrete stiffness tuning which enables robotic systems to behave as both traditional rigid linkages or as soft bodied continua. The price paid is a more limited workspace, but this, as we will show, is a design choice that can be made. Reiterating, trimmed helicoids allow a more expressive design space than using bulk materials, resulting in easily tunable stiffness properties based on geometric parameters. Combining both of these design methodologies allows one to have a great degree of control over the kinematic and structural properties of robot designs.


\subsection{Analysis}\label{sec:analysis}

\begin{figure}[ht]
\centering
\includegraphics[width=0.45\textwidth]{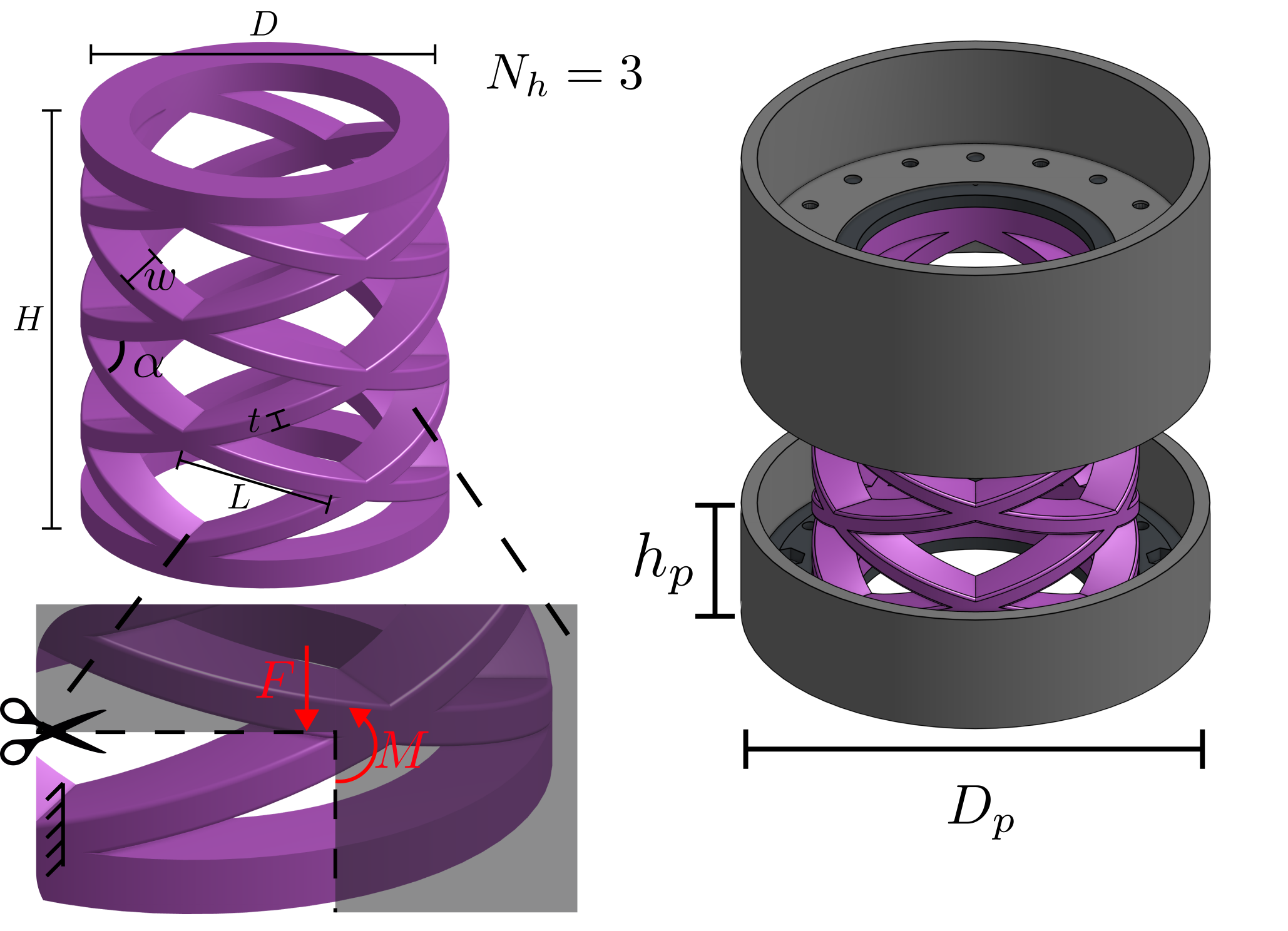}
\caption{Left Top: Important geometric parameters of the trimmed helicoid. Left Bottom: Free body diagram for our analysis. Right: kinematic parameters of soft-rigid components.}\label{fig:fbd}
\end{figure}

In the following, we discuss a new modeling methodology for THs that captures the structural mechanics in an elegant way. We will show how such modeling allows for closed-form design equations while also granting new insights into the behavior of these structures. The primary goal of our modeling approach is to approximate the structural stiffness of a trimmed helicoid in bending and compression modes of deformation. Our analysis hinges on the simple idea that the structure can be reasonably approximated by Euler-Bernoulli (EB) beam models, which relies on the assumption of small strain. This assumption may at first appear problematic; after all, our goal is to produce soft robots that can perform useful work through large deformations. However, as we will see, this architectured structure's geometry is such that the material strain in most cases is quite small ($< 5\%$), even as it allows large deformations of the bulk structure. 
We begin by defining terms according to the following definitions. Refer to Fig. \ref{fig:fbd} for a depiction of geometric variables. The base parameters of our trimmed helicoids will be height of the segment $H$, diameter $D$, strut width $w$, strut thickness $t$, and number of helices per segment $N_h$. As in \cite{guanTrimmedHelicoidsArchitectured2023a}, specifying these variables fully defines the geometry of the trimmed helicoid. We can also use these quantities to define additional geometric properties that will prove valuable in our analysis including the helical angle 
\begin{equation}\label{eqn:angle}
    \alpha = \arctan(\frac{2 H}{\pi D}),
\end{equation}
radius $R = D/2$, and a useful term $h = H/N_h$ that is the height over number of helices. Finally, we approximate the length of a single strut of the helicoid
\begin{equation}\label{eqn:len}
    L \approx \sqrt{(H/N_h)^2 + (\pi D /N_h)^2} / 2
\end{equation}

To confirm our hypothesis that the material strains our small, we examine the maximum material strain in a fully compressed structure. At fully compression, the maximum bending angle of a single strut of the helix is $\theta_{\mathrm{max}} = \alpha/2$. From elementary beam mechanics, we can then note that the maximum strain is experienced at the upper and lower edges of the strut (located at $\pm t/2$ in the vertical access of a cross section) and its value is therefore
\begin{equation}
    \epsilon_{\mathrm{max}} = \frac{t/2}{\rho} = \frac{t \alpha}{2 L},
\end{equation}
where $\rho$ is the curvature. Taking base values from \cite{guanTrimmedHelicoidsArchitectured2023a} ($H = 10$ cm, $D = 6$ cm, $N_h = 6$, $t = 3$ mm, $w = 6$ mm), we find for this case $\epsilon_{\mathrm{max}} = 4.58\%$. Assuming that this value is somewhat characteristic of the maximum strain across the practical design spectrum for these structures, EB beam theory will be a reasonable assumption as long as the model is used as a design heuristic (as opposed to, say, a safety critical structural analysis). 

We now present our analysis of the structural stiffness. We begin by reiterating our assumption of small strain. Similarly, we also assume a linear material model, and approximate the stress-strain behavior of silicone as such. We approximate the Young's modulus from the Shore A hardness using the famous Gent model \cite{gentRelationIndentationHardness1958a}, which has been shown to be highly accurate for liquid silicone rubbers (LSRs) \cite{larson2019can},
\begin{equation}
    E \approx \frac{0.0981 (56 + 7.62336 S)}{0.137505 (254 - 2.54 S)};
\end{equation}
where $E$ is the Young's modulus in MPa and $S$ is the Shore A hardness.

Critical to our following analysis is our idealization of the helices as beams. We consider that each strut forming the cells of our trimmed helicoid can be modeled as a statically indeterminate beam that is fixed at one end and free to translate (but not rotate) on the other end. Figure \ref{fig:fbd} shows this idealization. Using symmetry, we can easily derive the reaction forces and moments as follows 
\begin{gather}
    R_1 = F/2 \\
    M = M_1 = M_2 = \frac{1}{2}FL\cos(\alpha).
\end{gather}
To further simplify the analysis, we approximate the beams as straight given that they are typically short relative to the circumference of the bulk structure. Finally, we define the second moment of area using our previously defined variables as 
\begin{equation}
    I = \frac{1}{12}w t^3.
\end{equation}
It is now easy to solve the deflection problem using superposition of a cantilever with a concentrated force and a cantilever with a constant moment, which are standard textbook solutions \cite{budynas2011shigley}:
\begin{equation}\label{eqn:deflection}
    y = (\frac{FL^3}{3EI}\cos^2(\alpha) - \frac{ML^2}{2EI}\cos(\alpha)) = \frac{FL^3}{12EI}\cos^2(\alpha).
\end{equation}
For the particular loading situation, this deflection equation can be used to derive the approximate structural stiffness for a trimmed helicoid.

\subsubsection{Axial Stiffness}
In the compression analysis, we note that each the load is distributed over essentially $N_h$ sets of parallel springs, each consisting of $N_h$ series springs. Thus, finding the stiffness in compression for the whole structure is equivalent to finding that of a single strut. The critical salient feature of our compression model is that the values for the length of the beam and the angle of the loading condition and deflection vary across the radius of the helicoid, while our equations (\ref{eqn:len}) and (\ref{eqn:angle}) are taken at the outer edge of the cylinder. This is important because in compression the entire surface of each strut of the helicoid undergoes significant stress and strain. Therefore, we replace those nominal values with their counterparts at the center of each strut, or in other words at a radius of $r = (D - w)/2$,
\begin{gather}
    L_{\mathrm{avg}} \approx \frac{1}{2N_h}\sqrt{H^2 + (\pi (D - w) - 2t)^2} \\
    \alpha_{\mathrm{avg}} \approx \arctan(\frac{2 H}{\pi (D - w) - 2t}).
\end{gather}
Replacing the relevant terms in eqn (\ref{eqn:deflection}) gives us our final deflection
\begin{equation}
    y = \frac{FL_{\mathrm{avg}} ^3}{12EI}\cos^2(\alpha_{\mathrm{avg}}),
\end{equation}
from which we can trivially calculate stiffness as
\begin{equation}\label{eqn:axial}
    k_{\mathrm{ax}} = \frac{F}{y} = \frac{12EI}{L_{\mathrm{avg}} ^3\cos^2(\alpha_{\mathrm{avg}})}
\end{equation}

\begin{figure}[H]
\centering
\includegraphics[width=0.45\textwidth]{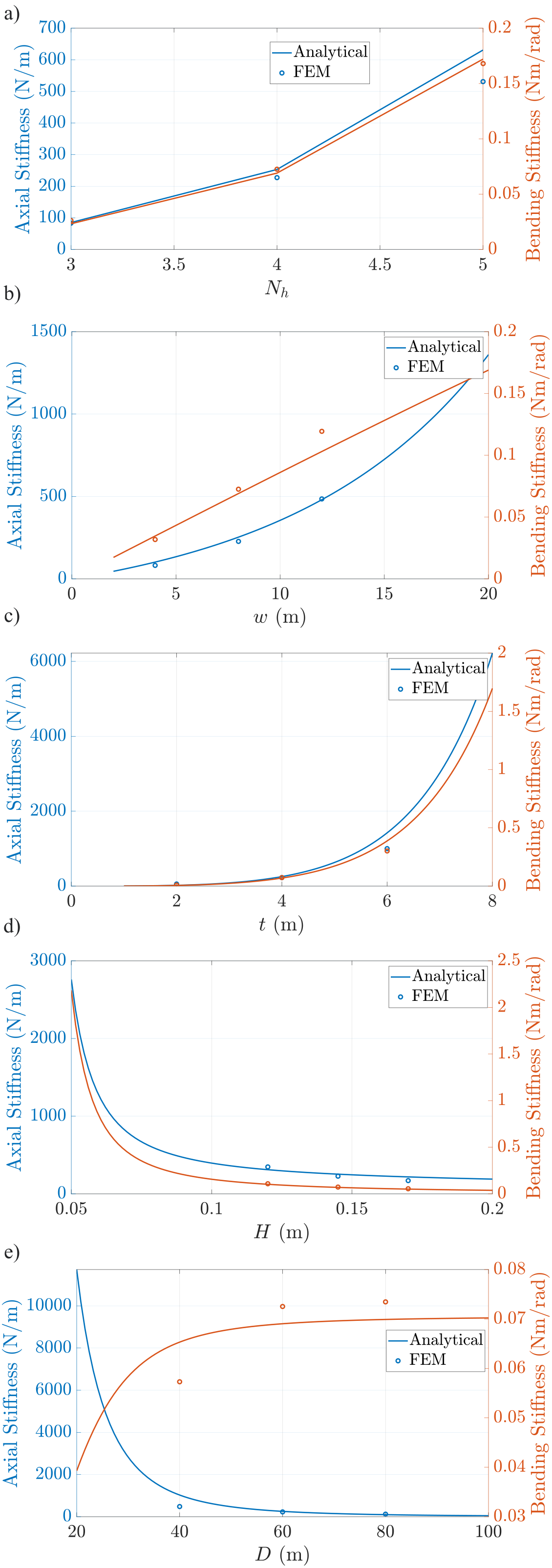}
\caption{Comparison of analytical model with FEM. a) Number of helices $N_h$. b) Radial width of each helix $w$. c) Thickness of each helix $t$. d) Height of the helix $H$, Diameter of the helix $D$.}\label{fig:analysis}
\end{figure}

We compare our analytical model with a Finite Element Method (FEM) analysis as a benchmark. To make the number of FEM analyses tractable, we vary each independent variable separately through three values and plot results in Fig. \ref{fig:analysis}.

\subsubsection{Bending Stiffness}
For our bending analysis, we use a rough approximation by first assuming that our helicoid is a rigid bar in tension and finding a stiffness of 
\begin{equation}
    k_{\mathrm{ax}} = \frac{EA}{H},
\end{equation}
where $A = \frac{\pi}{4} (2R - w)^2 = \pi R_m^2$.
For a solid bending beam
\begin{equation}
    k_{\mathrm{bend}} = \frac{EI}{H},
\end{equation}
where $I = \frac{\pi}{4} R_m^4$.
Thus, solving for $E$, we can write 
\begin{equation}
    k_{\mathrm{bend}} = \frac{k_{\mathrm{ax}}I}{A}.
\end{equation}
We found that this particular equation did not result in a good fit, but taking inspiration from the Smalley equation for wave springs \cite{smalley2014designing}, we found the following empirical adjustment resulted in a good fit
\begin{equation}
    k_{\mathrm{bend}} = \frac{9k_{\mathrm{ax}}I}{A}\frac{R_m}{H}.
\end{equation}

\subsubsection{Kinematics}
The kinematic workspace of the robot is determined by either the maximum deformation of the trimmed helicoid or the geometry of the rigid components. Here, we assume that the rigid components are designed to make contact before the trimmed helicoid (otherwise they do not have a structural role and only effect contact dynamics). The important kinematic parameters are the height of the rigid plate, $h_p$, and the diameter of the plate, $D_p$, which are shown in Figure \ref{fig:fbd}, as well as the number of intermediate plates in a segment, $N_p$ (i.e. not the end plates, so in Figure \ref{fig:fbd} we have $N_p = 0$). Maximum compression is determined purely by $h_p$ as 
\begin{equation}
    \delta L_{\mathrm{max}} = H - 2h_p(N_p + 1).
\end{equation}
Maximum bending is estimated with elementary constant curvature kinematics as 
\begin{equation}
    \theta_{\mathrm{max}} \approx \frac{2\delta L}{D_p}.
\end{equation}

\subsection{Design Tool}\label{sec:dt}
A core goal of this work was to produce a flexible parametric design tool that could be used to create arbitrary instantiations of our design by simply varying a limited and intuitive set of design variables. Users can simply update the desired parameters and a new model will automatically generate, including the necessary mold parts (discussed in more detail in the next section). The CAD was completed in Onshape and is available to the community through our project's GitHub repository \footnote{\url{https://github.com/zpatty/sr_helix}}. Readers who wish to create their own design can make a copy of the Onshape project and modify the relevant design parameters (located at the top of the feature tree under the Variables folder). The model will then automatically regenerate.

\section{Manufacturing}\label{sec:manufacturing}
While the trimmed helicoid concept was initially prototyped using direct 3D printing of printable thermoplastic elastomers (TPUs) \cite{guanTrimmedHelicoidsArchitectured2023a}, injection molding represents an attractive alternative manufacturing procedure for producing the parts. By molding, we can choose from a wider and more affordable variety of commercial and industrial grade elastomer products including genuine silicone rubbers. While great progress has been made in the last five years in printing materials with elastomeric qualities, most 3D printed elastomers still fall short of their manufacturing-grade counterparts in one material property or another. They are also quite expensive. Finally, while 3D printing is attractive for rapid prototyping in the lab, injection molding provides a direct avenue to scale up manufacturing arbitrarily. Additionally, the injection molds themselves can be 3D printed and thus this approach still maintains similar design-to-prototype speeds. 

\begin{figure}[h]
\centering
\includegraphics[width=0.4\textwidth]{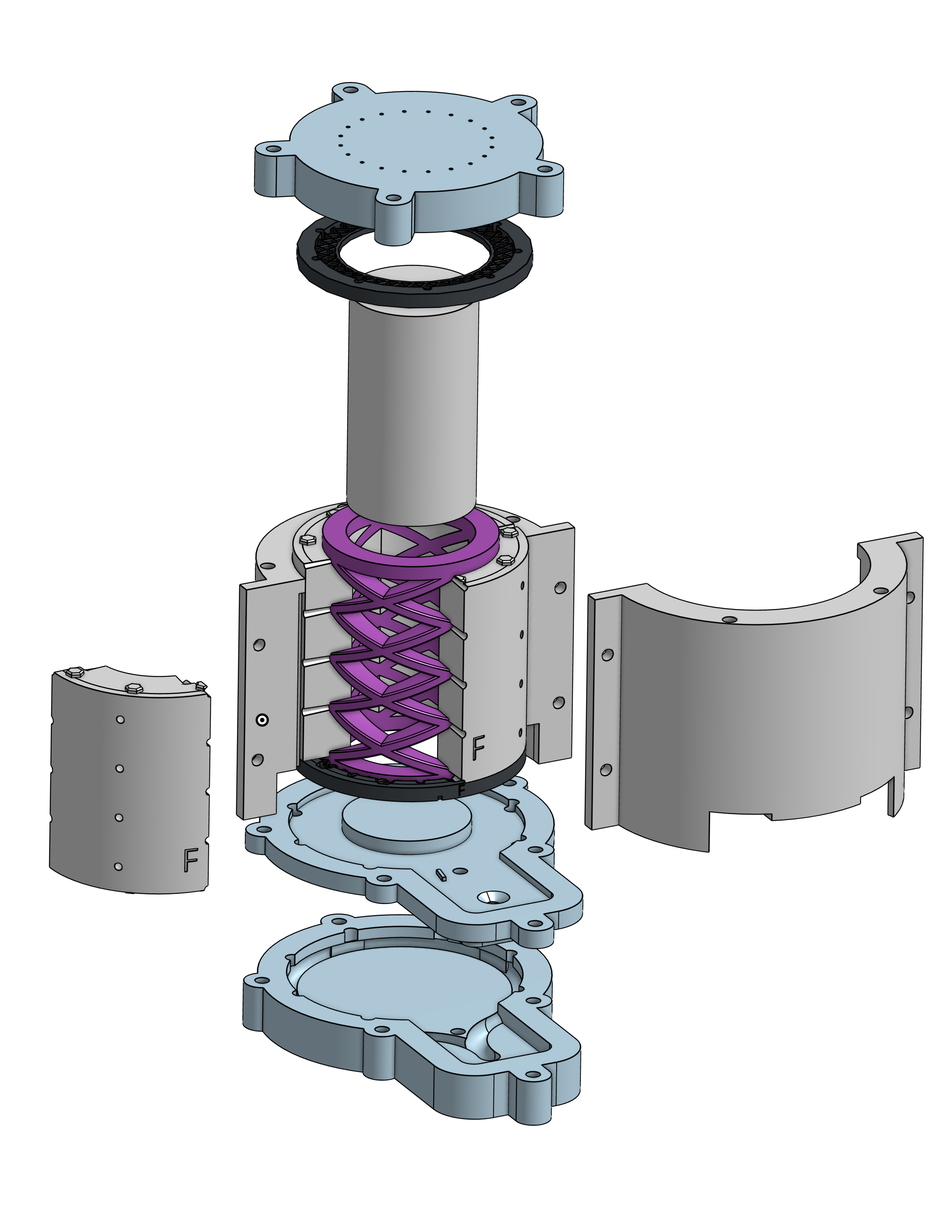}
\caption{Exploded view of the mold.}\label{fig:mold}
\end{figure}

We utilize an injection molding setup based on \cite{bellInjectionMoldingSoft2022}, wherein we inject liquid silicone rubber at room temperature into a 3D printed mold with a 2-part pneumatic caulking gun. The silicone is automatically mixed in a standard static mixing nozzle. The mold can then be left at room temperature to cure, or else placed in an oven at an appropriate temperature (subject to the thermal properties of the particular 3D printed material) to accelerate the curing process. In this work, we utilized Bambu P1S PLA for our molds, allowing us to take advantage of the very high print speed of these machines in order to rapidly prototype. 

Our mold consists of $6+N_h$ parts and facilitates simple assembly and demolding. The struts of the helicoid are slightly drafted to allow easier release, and the part that contains the actual mold cavities is cut (in CAD) into $N_h$ (the number of helices). In our prototyping phase, we found that using one mold cavity part per helix interface was the key to eliminating part destruction during the demolding step. Although this is a relatively complex part, thanks to the regular structure only a single core is necessary to mold it, which results in easy assembly and disassembly. All other parts provide features such as the sprue, air vents (to prevent air bubbles), clamping and fastening, etc. An exploded view of a characteristic mold is provided in Figure \ref{fig:mold}. As discussed in Section \ref{sec:dt}, we provide a CAD tool to automatically generate molds for any such prototype. Mold parts can then be exported for printing or machining.

\section{Experimental Characterization}
\begin{figure}
\includegraphics[width=0.48\textwidth]{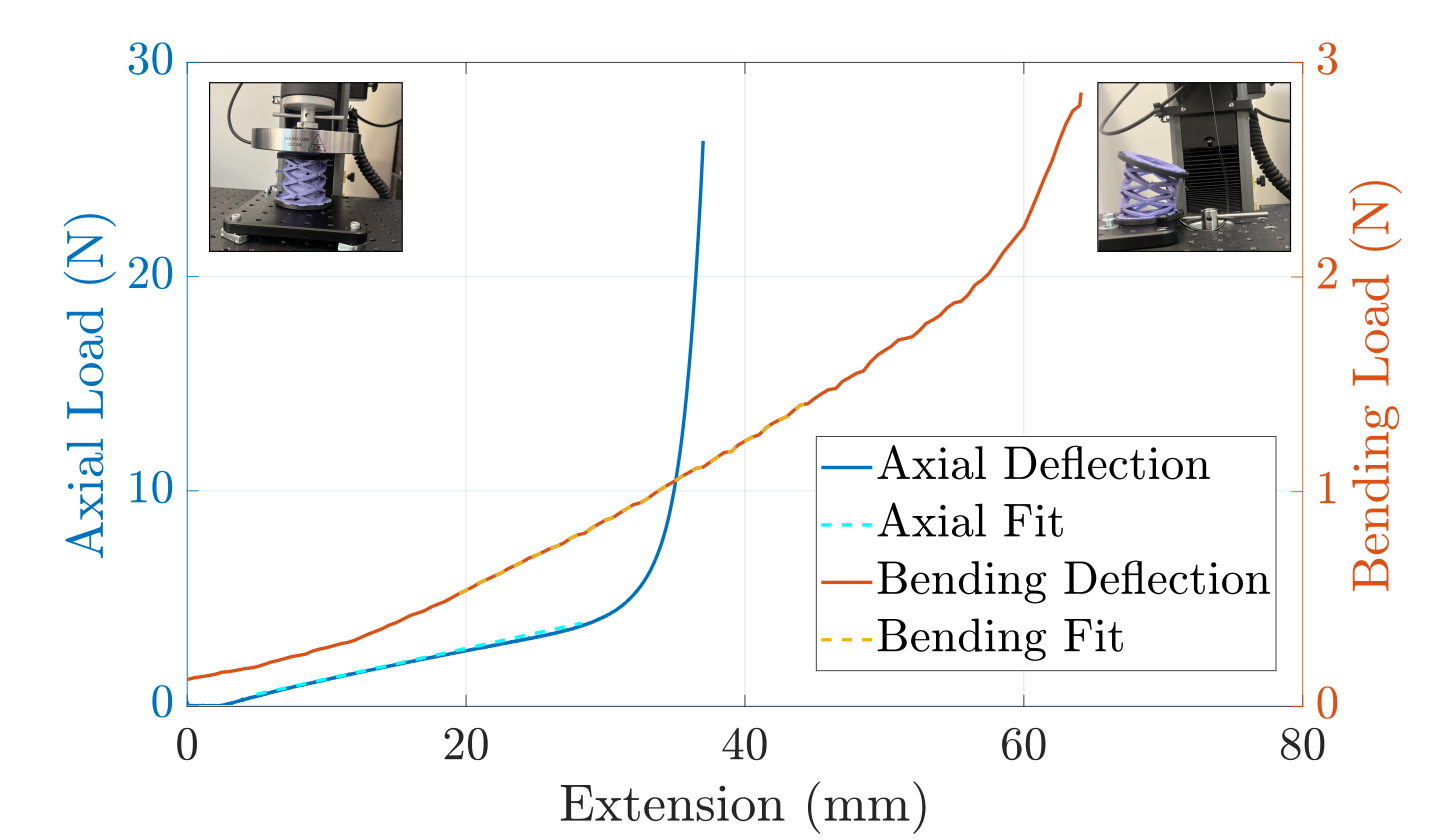}
\caption{Example plots of characterization experiments for compression and bending stiffness. The procedure for axial compression is inset at the top left and the procedure for bending at the top right.}\label{fig:characterization}
\end{figure}

We characterize prototypes using an Instron materials testing system. For axial experiments, we directly compress the module and find the slope of the force-displacement curve. For bending experiments, we attach a cable to the top plate of the module and pull it through a simple transmission with the Instron, measuring tension and bending displacement. Depictions of these experiments are shown inset in Fig. \ref{fig:characterization}. We find that in both compression and bending, the helicoid force-displacement response is linear (until struts begin to make contact). Characteristic plots of the force-displacement curves are shown in Fig. \ref{fig:characterization}. We compare for two module designs. The first, which we call the "Small and flexible" module is defined by setting height of the segment $H = 0.12 \mathrm{m}$, diameter $D = 0.06 \mathrm{m}$, strut width $w = 0.008 \mathrm{m}$, strut thickness $t = 0.004 \mathrm{m}$, and number of helices per segment $N_h = 3$. The second, which we call the "Large and stiff" module is defined by setting height of the segment $H = 0.17 \mathrm{m}$, diameter $D = 0.08 \mathrm{m}$, strut width $w = 0.012 \mathrm{m}$, strut thickness $t = 0.005 \mathrm{m}$, and number of helices per segment $N_h = 4$. We tested 6 samples of the small design and a single sample of the large design. Results for our experiments are shown in Table \ref{tab:results}. Percent error for the analytical model relative to the physical characterization ranges from $1\%$ to $20\%$.

\begin{table}[t]
\centering

\vspace{5mm}

\caption{Comparison of experimental data, analytical model, and FEM analysis for two module designs.}
\label{tab:results}
\resizebox{0.46\textwidth}{!}{
\begin{tabular}{lccc}
\toprule
\textbf{Module Design} & \textbf{Method} & \textbf{Axial Stiffness ($\frac{\mathrm{N}}{\mathrm{m}}$)} & \textbf{Bending Stiffness ($\frac{\mathrm{Nm}}{\mathrm{rad}}$)}\\ 
\midrule
\multirow{3}{*}{Small and flexible} 
 & Experiment & $124.4 \pm 11.1$ & $0.0321 \pm 0.006$ \\
 & Analytical & $99.0$ & $0.0326$ \\
 & FEM & $112.4$ &  $0.0223$ \\
\midrule
\multirow{3}{*}{Large and stiff} 
 & Experiment & $410.2$ & $0.1452$ \\
 & Analytical & $354.6$ & $0.185$ \\
 & FEM & $393.1$ & $0.130$ \\
 
\bottomrule
\end{tabular}}
\end{table}
\section{Robot}
\subsection{Design}
\begin{figure*}
\centering
\includegraphics[width=0.85\textwidth]{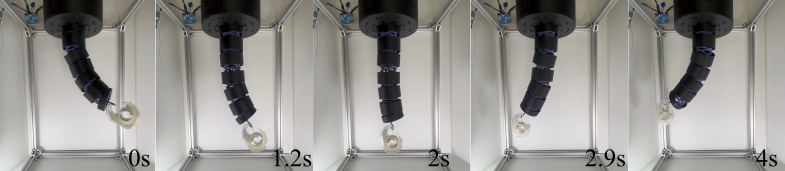}
\caption{Time lapse of a robot experiment.}\label{fig:tape}
\end{figure*}

We assemble multiple segments of our auto-generated trimmed helicoid structure to produce a serially connected soft-rigid manipulator. The helicoid design chosen is one of the more flexible ones tested, with parameters $H = 0.12 \mathrm{m}$, diameter $D = 0.06 \mathrm{m}$, strut width $w = 0.008 \mathrm{m}$, strut thickness $t = 0.004 \mathrm{m}$, and number of helices per segment $N_h = 3$. The robot has three modules, where each module is composed of two helicoid segments. A rigid plate is screwed on between each helicoid segment. There are 10 XM430-W350 Dynamixels housed in the rotating base. One motor is for rotating the robot in the z-axis, and the other 9 are attached to tendons. Each module is controlled by three tendons. To account for friction, we route the tendons for the second and third module through bowden tubes that run through the inside of the manipulator. The weight of the robot manipulator itself is about $630$ g, while the combined weight of the motor assembly is about $1$ kg.

\subsection{Robot Controlled Experiments}
\begin{figure}
\includegraphics[width=0.48\textwidth]{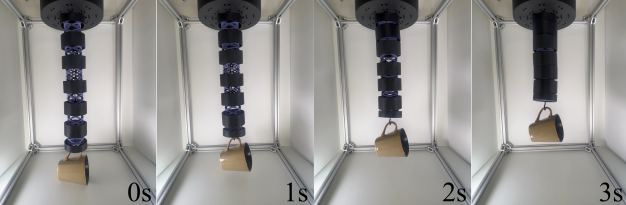}
\caption{Robot lifting a 400 gram mug. For scale, at full extension the robot is approximately 0.45 meters.}\label{fig:mug}
\end{figure}

We conducted several experiments to show the capabilities of our system with respect to compliance and payload, the key benefits of the soft-rigid design. We approximate the kinematics using the Piecewise Constant Curvature (PCC) assumption \cite{dellasantinaImprovedStateParametrization2020a} and we calculate the state in that parametrization based on the encoder-derived cable lengths of the tendons. We implemented purely kinematic control based on low-level position control of the motors, as well as basic PD control based on low level torque control with elementary feedforward to account for some of the dynamics. We do not report tracking experiments as the development of the novel tracking controller is beyond the scope of this paper. Snapshots of a few of the experiments we conducted are shown in Figure \ref{fig:tape} and \ref{fig:mug}. In Figure \ref{fig:tape}, we demonstrate the robot manipulating a light object through some of the robot's range of motion, and in Figure \ref{fig:mug}, we show the robot lifting a $400$ gram mug. Both experiments use the aforementioned current torque loop. We show additional experiments in the accompanying video.
\section{Discussion}

In summary, we presented a new robot design paradigm which combines previous innovations in soft-rigid hybrid robots and trimmed helicoid architectured structures. We contributed an open source design and manufacturing pipeline capable of producing production-quality soft trimmed helicoids through injection molding. We also developed a new analytical model that can be used as a design tool to predict the structural properties of trimmed helicoids based on geometric properties. Finally, we assembled a proof of concept robot using the above contributions. 

There are several avenues to build off of this research. First, our analytic model could be improved. In this work, we sought to maintain the utmost simplicity for easy interpretability and extension. However, the trimmed helicoid structure is complex and our models may benefit from introducing additional considerations, such as modeling the beams as curved \cite{young2002roark}. Additionally, while investigating potential models for bending, we found that the most obvious purely analytic extensions did a good job of modeling the effect of our parameters $w$, $t$, and $N_h$, but did a poor job of modeling the effect of the diameter $D$ and the height $H$. This is why we ultimately chose to use a semi-empirical model. As such, there is an open question as to whether a straightforward first-principles model can be derived for trimmed helicoid bending. Nonetheless, we have found our model useful as a first step to guide design intuition. 

The robot produced in this work uses very flexible trimmed helicoids, making its whole body flexible. This was an intentional choice, as we observed that many soft robot manipulators in the literature are in practice quite stiff, and thus we wanted to create a very soft robot that was nonetheless capable of relatively high payload (in this case, we demonstrated our robot lifting $63\%$ of the mass of the manipulator or $28\%$ of the mass of the manipulator plus motor apparatus). However, increasing the stiffness of the manipulator would likely result in improved performance without sacrificing the kinematic and interaction benefits of overall structural deformability. Additionally, our control algorithms and models were very basic. Given the relatively low stiffness of the robot, more sophisticated modeling and control is necessary to properly drive the system \cite{dellasantinaModelBasedControlSoft2023}.

Finally, the robot uses cable lengths to perform state estimation, as is common in the tendon-driven soft robot literature \cite{guanTrimmedHelicoidsArchitectured2023a}. This is at best a crude estimation of the robot's state, and performance would benefit from additional sensing methodologies for more comprehensive state estimation. Due to the structural regularity of the trimmed helicoid, it is likely that relatively sparse sensing could dramatically improve state estimation. Future research on this topic should seek to tightly integrate sensing into the robot structure.



\bibliographystyle{ieeetran}
\bibliography{Robosoft2025}

\end{document}